\documentclass{article} 
\usepackage{iclr2025_conference,times}

\usepackage[utf8]{inputenc} 
\usepackage[T1]{fontenc}    
\usepackage{hyperref}       
\usepackage{url}            
\usepackage{booktabs}       
\usepackage{amsfonts}       
\usepackage{nicefrac}       
\usepackage{microtype}      
\PassOptionsToPackage{numbers}{natbib}
\usepackage{natbib}

\usepackage{graphicx}
\usepackage{hyperref}
\usepackage{subcaption}
\usepackage{caption}
\usepackage{booktabs}
\setcitestyle{square}

\usepackage{booktabs}
\usepackage{amsmath}
\usepackage{makecell} 
\usepackage{hyperref}
\usepackage{graphicx}
\usepackage{subcaption}

\usepackage{amsmath,amsfonts,bm}

\def\eqref#1{equation~\ref{#1}}

\def\1{\bm{1}}

\DeclareMathAlphabet{\mathsfit}{\encodingdefault}{\sfdefault}{m}{sl}
\SetMathAlphabet{\mathsfit}{bold}{\encodingdefault}{\sfdefault}{bx}{n}

\usepackage{hyperref}
\usepackage{url}

\title{Multivariate LSTM-Based Forecasting for Renewable Energy: Enhancing Climate Change Mitigation}

\author{%
Farshid~Kamrani\\
  \shortstack[l]{Department of Mechanical and Aerospace \\Engineering}\\
  Carleton University\\
  Ottawa, ON, Canada\\
  \texttt{FarshidKamrani@cmail.carleton.ca}
  \And
  Kristen~Schell\\
  \shortstack[l]{Department of Mechanical and Aerospace \\Engineering}\\
  Carleton University\\
  Ottawa, ON, Canada\\
  \texttt{KristenSchell@cunet.carleton.ca}
}

\iclrfinalcopy 
\begin{document}

\maketitle

\begin{abstract}
The increasing integration of renewable energy sources (RESs) into modern power systems presents significant opportunities but also notable challenges, primarily due to the inherent variability of RES generation. Accurate forecasting of RES generation is crucial for maintaining the reliability, stability, and economic efficiency of power system operations. Traditional approaches, such as deterministic methods and stochastic programming, frequently depend on representative scenarios generated through clustering techniques like K-means. However, these methods may fail to fully capture the complex temporal dependencies and non-linear patterns within RES data. This paper introduces a multivariate Long Short-Term Memory (LSTM)-based network designed to forecast RESs generation using their real-world historical data. The proposed model effectively captures long-term dependencies and interactions between different RESs, utilizing historical data from both local and neighboring areas to enhance predictive accuracy. In the case study, we showed that the proposed forecasting approach results in lower $CO_2$ emissions, and a more reliable supply of electric loads.

\end{abstract}

\section{Introduction}

Renewable energy sources (RESs) have garnered significant attention in recent years due to their environmentally friendly nature and declining costs \cite{hassan2024renewable}. RESs, such as solar and wind power, play a critical role in modern power systems \cite{kamrani2019investigating}. However, their inherent variability poses considerable challenges to the reliable operation of these systems. Accurate energy generation forecasting is essential for balancing power supply and demand and maintaining grid stability \cite{kamrani2021flexibility}. Traditional methods for addressing these challenges often rely on either deterministic approaches \cite{kamrani2019investigating}\cite{kamrani2021flexibility} or stochastic programming \cite{kamrani2021two}. In the literature, K-means clustering has been widely employed to reduce large historical datasets into a smaller set of representative days that capture the uncertainty inherent in RESs. These representative days can serve as scenarios for stochastic programming, enabling more robust decision-making in power system operations \cite{kamrani2021two}. 

With the advancement of Machine Learning (ML), its applications have expanded across a wide range of disciplines. Authors in \cite{yang2017towards} proposed a new joint dimensionality reduction (DR) and K-means clustering approach to recover the ‘clustering-friendly’ latent representations and to better cluster the data. Recurrent Neural Networks (RNNs), particularly suited for sequential data, are commonly used to capture latent features and predict future time intervals. These models have proven effective in various time series prediction tasks \cite{waqas2024critical}. 
The authors in \cite{elsworth2020time} proposed combining recurrent neural networks with a dimension-reducing symbolic representation for time series forecasting. This approach addresses issues like sensitivity to hyperparameters and random weight initialization, while enabling faster training without sacrificing forecasting performance. The authors in \cite{abbasimehr2022improving} proposed a hybrid model combining LSTM and multi-head attention for accurate time series forecasting.

Holland et al. \cite{holland2022marginal} examined why marginal carbon emissions in the U.S. electricity sector have not declined despite increasing renewable energy adoption. The study estimates marginal emissions across different regions and time periods, finding that shifts in the energy mix, particularly the reliance on natural gas rather than renewables for marginal generation, play a significant role. The authors discuss the implications of these findings for climate policy, suggesting that carbon pricing and more targeted policies are needed to ensure emissions reductions. Building on these findings, accurate forecasts of RESs generation are essential for addressing this challenge. Reliable generation forecasts enable better grid management, enhance the integration of renewables, and reduce reliance on carbon-intensive generation.

The authors in \cite{effenberger2024towards} proposed a method for multi-decadal wind power forecasting that incorporates turbine locations into Gaussian Process-based downscaling, enabling accurate aggregate wind power predictions from the low-resolution CMIP6 climate model data.
The authors in \cite{ashhab2024feasibility} investigated the application of deep learning techniques for forecasting noisy and high-resolution power grid frequency time series, highlighting the effectiveness of the Temporal Fusion Transformer (TFT). In the context of power systems, RESs may be correlated with other types of renewable energies or with RESs at different locations \cite{Correlation}. 

The primary contribution of this paper is the development of a Multivariate Long Short-Term Memory (M-LSTM)-based network for forecasting RESs generation. The proposed model leverages real-world data of RESs from various locations in Alberta province. Notably, the model predicts RES generation for a specific area by utilizing its historical data as well as data from neighboring areas. We also consider the temporal gap
between the day-ahead market closure and the start of the real-time market in the
forecasting of RES. This contribution is pivotal for advancing RESs utilization and supporting climate change mitigation efforts.

\section{Proposed Model}
We propose a multivariate LSTM network designed to capture latent features and long-term dependencies among various RESs across different regions. By leveraging historical data, the model aims to enhance forecasting accuracy. The following sections provide a detailed explanation of the proposed approach.
\subsection{Structure of Network}
The proposed model employs a two-layer LSTM network. The first layer consists of 64 LSTM units with a Rectified Linear Unit (ReLU) activation function, while the second layer includes 32 LSTM units, also with ReLU activation. The outputs from the second LSTM layer are passed through a Dropout layer to prevent overfitting and then fed into a fully connected Neural Network for final predictions.


\subsection{Training the Multivariate LSTM}

\begin{figure}
    \centering
    \begin{subfigure}[b]{0.4\textwidth}
        \centering
        \includegraphics[width=\textwidth]{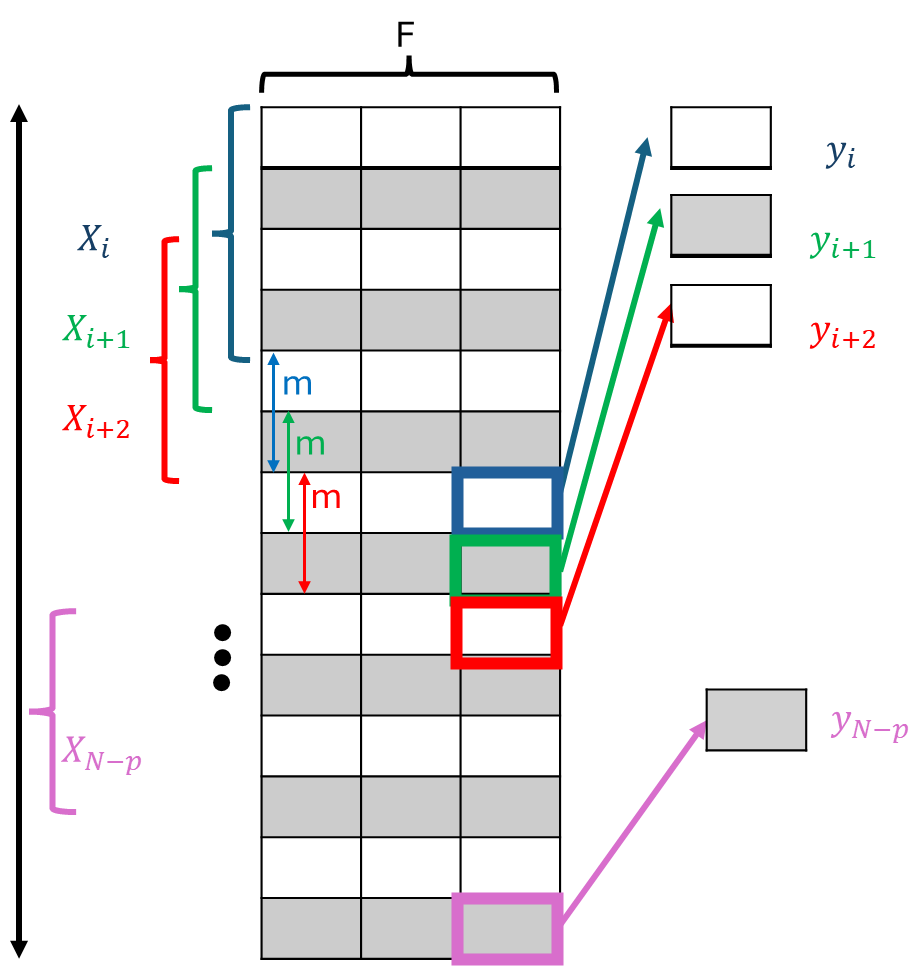}
        \caption{Training Structure}
        \label{fig:Training2}
    \end{subfigure}
    \hspace{0.05cm}
    \begin{subfigure}[b]{0.22\textwidth}
        \centering
        \includegraphics[width=\textwidth]{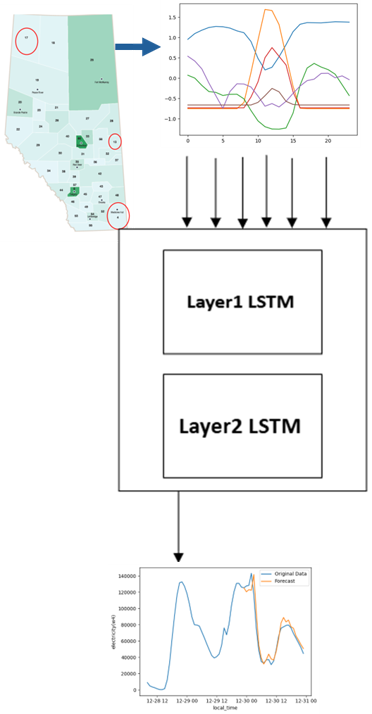}
        \caption{ Input/Output}
        \label{fig:InOut1}
    \end{subfigure}
    \caption{The Proposed M-LSTM }
    \label{fig:twofigs}
\end{figure}

Given a dataset of $N$ samples $x_{i,j}$, where $i = 1, \ldots, N$ (time steps) and $j = 1, \ldots, F$ (features), we define the inputs and outputs as follows:
The first input consists of a sequence of data points, $x_{i,j}$, from $i=1$ to $i=p$ (the number of look-back steps) for all $j$. The second input is a shifted sequence from $i=2$ to $i=p+1$ for all $j$. Correspondingly, the first output label is $x_{p+m,J}$, and the second output label is $x_{p+m+1,J}$, where $J$ represents the feature to predict. These outputs represent predictions, with $m$ steps ahead of the inputs. This configuration is used to define multiple samples and their corresponding labels throughout the dataset. Fig \ref{fig:Training2} illustrates the data samples and labels in the training process.


To ensure effective training, the input data was normalized, addressing potential issues caused by unscaled features. Normalization mitigates the sensitivity of activation functions to input scale and prevents large-scale features from dominating the learning process at the expense of smaller-scale features.
 The normalized data was fed into the network, which used the Adam optimizer for weight updates and Mean Squared Error (MSE) as the loss function. After training, the network's outputs were rescaled to their original scale. 
Also, we incorporated prior knowledge regarding periods when solar PV systems generate no power. Specifically, for predefined dark hours (i.e., nighttime), we explicitly set the PV generation to zero. This approach eliminates the need for predictions during these hours, ensuring that the model's outputs remain both accurate and consistent with real-world conditions.

\subsection{ Experiments}
We used real renewable energy generation historical data from the Alberta province \footnote{Available at: \url{https://www.aeso.ca/market/market-and-system-reporting/data-requests/historical-generation-data}}, which includes a full year of data, at one hour time intervals. We considered one type of RES - PV - across three different planning areas in the Alberta province. Our historical dataset consists of 8760 data samples per feature, with three features in total. The model is designed to look back at the past 24 hours of generation data for PV generation across all regions and predict the generation of the desired renewable source for the target time step. Fig \ref{fig:InOut1} illustrates the input and output of our proposed model.

\section{ Results}

\begin{figure}[ht]
  \centering
  \includegraphics[width=0.45\textwidth]{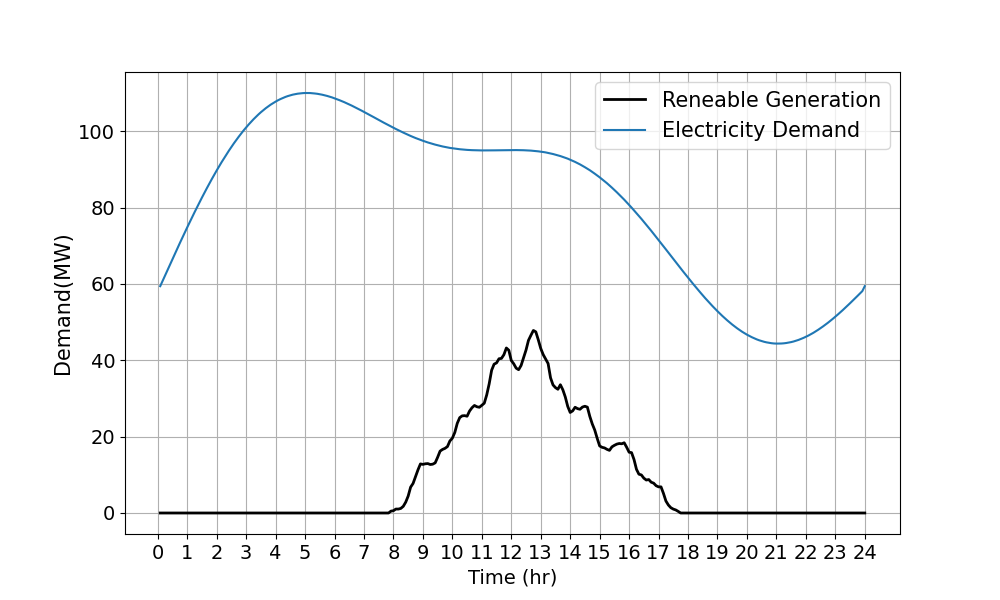}
  \caption{Demand to be supplied, and Actual PV Generation}
  \label{fig:Demand}
\end{figure}

To quantify the impact of improved RES forecasting on reducing greenhouse gas emissions, we consider an Economic Dispatch (ED) problem. Three types of synchronous generators and one PV source are available to meet demand. Table\ref{table:Generators} and Fig\ref{fig:Demand} provide a detailed overview of this system. The forecasted RES generation is utilized in the day-ahead (DA) ED. However, in the real-time (RT) ED, the actual RES generation values become available and are used to adjust the system operation accordingly. The day ahead and real-time ED formulations are provided in the \ref{DA} and \ref{RT} respectively.

Based on the DA operation and the RT actual RES generation, the real-time ED must be run to adjust the output of the generators. The discrepancy between the forecasted and actual RES values must be supplied by conventional generators in real-time. These generators are highly flexible, typically powered by natural gas, and contribute to $CO_2$ emissions. Based on data from  \cite{C2ES2015} and \cite{EEAGrants}, $CO_2$ emissions for this type of generator fall within a specific range. In this paper, we assumed 202 kg $CO_2$ per MWh for these generators.
If the available generator capacity is insufficient to cover this discrepancy, load shedding may occur, which is costly. Conversely, if the actual RES generation exceeds the forecasted values, it can either help reduce the need for conventional generators and load shedding from the day-ahead operation, or be spilled if it cannot be fully utilized.

\begin{figure}
    \centering
    \begin{subfigure}[b]{0.3\textwidth}
        \centering
        \includegraphics[width=\textwidth]{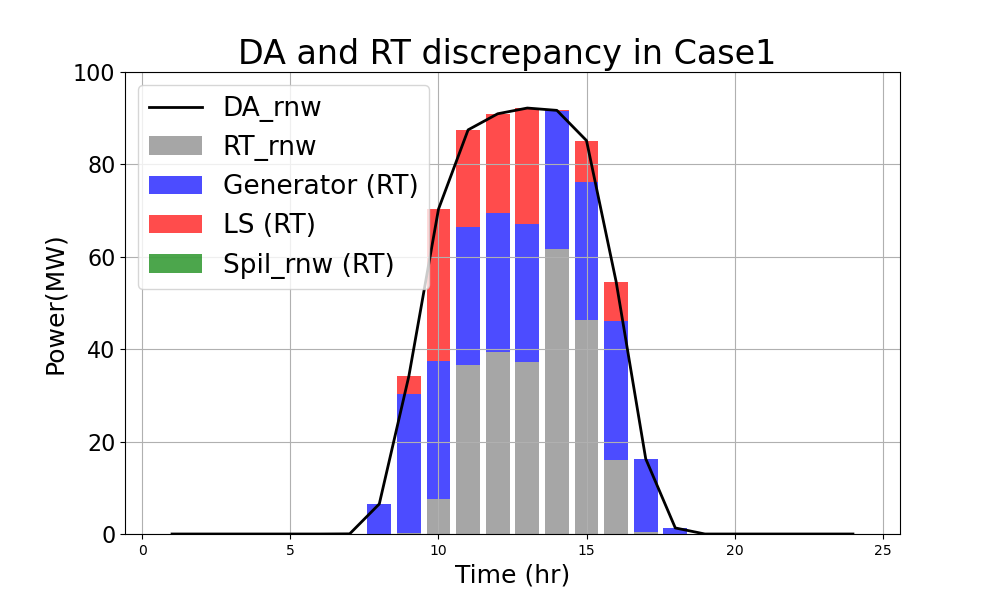}
        \caption{Case 1}
        \label{fig:subfig1}
    \end{subfigure}
    \hfill
    \begin{subfigure}[b]{0.33\textwidth}
        \centering
        \includegraphics[width=\textwidth]{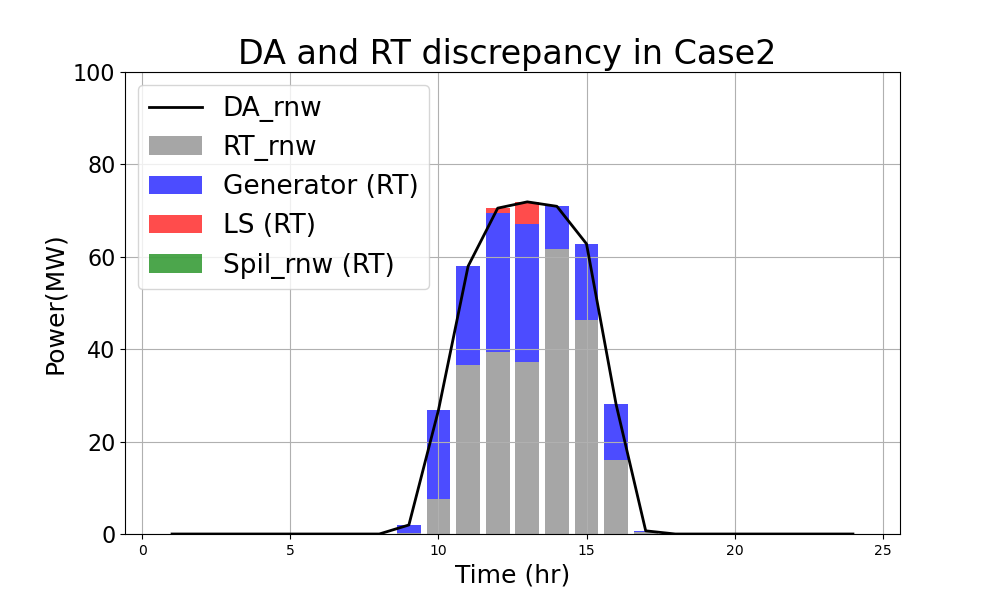}
        \caption{Case 2}
        \label{fig:subfig2}
    \end{subfigure}
    \hfill
    \begin{subfigure}[b]{0.33\textwidth}
        \centering
        \includegraphics[width=\textwidth]{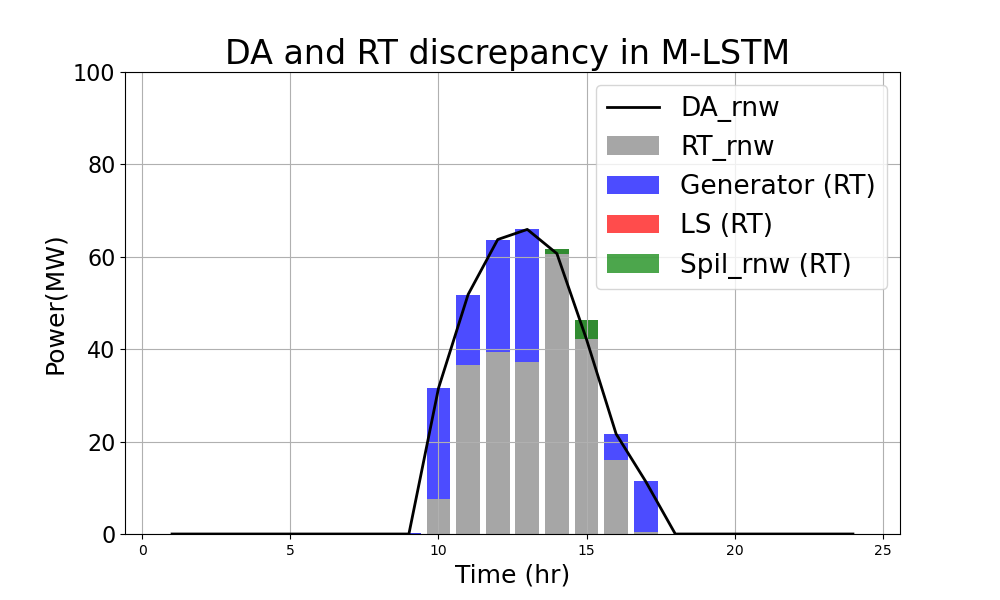}
        \caption{Case 3}
        \label{fig:subfig3}
    \end{subfigure}
    \caption{DA and RT discrepancy in three cases}
    \label{fig:threefigs}
\end{figure}

We compare three models for forecasting the PV generation in DA market: (1) a K-means approach \cite{kamrani2021two}, (2) a monthly  average at each hour, and (3) our M-LSTM model. The results show the deviation between the DA market and the RT market based on the actual values of RES. Table \ref{tab:comparison} and Fig. \ref{fig:threefigs} illustrate the results of ED in the DA and RT markets with three different forecast of PV.

\begin{table}[h]
    \centering
    \begin{tabular}{lccc}
        \toprule
        & Case1  & Case2 &M-LSTM\\
        \midrule
        Gas fired unit (MW) & 263.7  & 140.56&108.7 \\
        $CO_2$ emission (kg) & 53,267 &28,393 &21,957 \\
        Load Shedding (MW) & 121.7 & 5.79 &0 \\
        RES Spillage (MW) & 0 & 0 &5 \\
        DA+RT Cost (\$)& 53,043 & 44,148&43,490\\
        NMAE &1.57&0.59 &0.46\\
        \bottomrule
    \end{tabular}
    \caption{Model Comparison and impact on power grid metrics}
    \label{tab:comparison}
\end{table}

The dramatic reduction in gas-fired unit usage from 263.7 MW in Case 1 to 108.7 MW with M-LSTM highlights the potential for significant reduction in fossil fuel use in power systems through advanced AI-driven management techniques. This also substantially decreases $CO_2$ emissions across the different cases, with the M-LSTM approach achieving the lowest emissions at 21,957 kg, compared to 53,267 kg in Case 1 - a remarkable 58\text{\%} decrease.
The M-LSTM method shows superior performance in load management, completely eliminating load shedding (0 MW) compared to 121.7 MW in Case 1 and 5.79 MW in Case 2, indicating its potential for enhancing grid stability and reliability. 
M-LSTM's slightly higher spillage reflects its superior forecast accuracy. Other models avoid spillage by overestimating renewable generation, which masks forecast errors. For M-LSTM, spillage only occurs when actual real-time generation surpasses accurate forecasts after meeting all demand and minimizing conventional generation. This minimal spillage demonstrates optimal system balancing, maximizing renewable usage while maintaining grid stability and reducing reliance on conventional power. The approach effectively balances the trade-off between renewable integration and system reliability.
Based on the Normalized Mean Absolute Error (NMAE) using Mean Normalization, M-LSTM exhibits the lowest error, indicating the highest prediction accuracy compared to the ground truth.
These results underscore the potential of machine learning approaches, particularly M-LSTM, in optimizing power system operations to simultaneously reduce carbon emissions, minimize load shedding, and efficiently integrate renewable energy resources. Additionally, the cost of ED in Case 3 is lower than in the other two cases, suggesting that more accurate forecasting leads to better system performance in terms of both economics and $CO_2$ emissions.

\section{ Conclusion}
The proposed M-LSTM model enhances renewable energy forecasting by capturing complex temporal and spatial dependencies across multiple locations and RES types using historical data. Accurate forecasts lead to more efficient economic dispatch, reduced operational costs, enhanced system reliability, and reducing $CO_2$ emissions by 58\%. This research highlights the critical role of advanced machine learning techniques in improving renewable energy utilization in modern power systems, to mitigate climate change. Results demonstrate that the scaled use of the M-LSTM model across power generation systems could significantly reduce $CO_2$ emissions.

\bibliographystyle{plainnat} 
\bibliography{iclr2025_conference} 

@article{kamrani2021flexibility,
  title={Flexibility-based operational management of a microgrid considering interaction with gas grid},
  author={Kamrani, Farshid and Fattaheian-Dehkordi, Sajjad and Abbaspour, Ali and Fotuhi-Firuzabad, Mahmud and Lehtonen, Matti},
  journal={IET Generation, Transmission \& Distribution},
  volume={15},
  number={19},
  pages={2673--2683},
  year={2021},
  publisher={Wiley Online Library}
}

@article{kamrani2021two,
  title={A two-stage flexibility-oriented stochastic energy management strategy for multi-microgrids considering interaction with gas grid},
  author={Kamrani, Farshid and Fattaheian-Dehkordi, Sajjad and Gholami, Mohammad and Abbaspour, Ali and Fotuhi-Firuzabad, Mahmud and Lehtonen, Matti},
  journal={IEEE Transactions on Engineering Management},
  volume={70},
  number={10},
  pages={3330--3343},
  year={2021},
  publisher={IEEE}
}

@inproceedings{kamrani2019investigating,
  title={Investigating the impacts of microgrids and gas grid interconnection on power grid flexibility},
  author={Kamrani, Farshid and Fattaheian-Dehkordi, Sajjad and Abbaspour, Ali and Fotuhi-Firuzabad, Mahmud and Lehtonen, Matti},
  booktitle={2019 smart grid conference (SGC)},
  pages={1--6},
  year={2019},
  organization={IEEE}
}

@inproceedings{yang2017towards,
  title={Towards k-means-friendly spaces: Simultaneous deep learning and clustering},
  author={Yang, Bo and Fu, Xiao and Sidiropoulos, Nicholas D and Hong, Mingyi},
  booktitle={international conference on machine learning},
  pages={3861--3870},
  year={2017},
  organization={PMLR}
}

@article{waqas2024critical,
  title={A critical review of RNN and LSTM variants in hydrological time series predictions},
  author={Waqas, Muhammad and Humphries, Usa Wannasingha},
  journal={MethodsX},
  pages={102946},
  year={2024},
  publisher={Elsevier}
}

@article{hassan2024renewable,
  title={The renewable energy role in the global energy Transformations},
  author={Hassan, Qusay and Viktor, Patrik and Al-Musawi, Tariq J and Ali, Bashar Mahmood and Algburi, Sameer and Alzoubi, Haitham M and Al-Jiboory, Ali Khudhair and Sameen, Aws Zuhair and Salman, Hayder M and Jaszczur, Marek},
  journal={Renewable Energy Focus},
  volume={48},
  pages={100545},
  year={2024},
  publisher={Elsevier}
}

@article{elsworth2020time,
  title={Time series forecasting using LSTM networks: A symbolic approach},
  author={Elsworth, Steven and G{\"u}ttel, Stefan},
  journal={arXiv preprint arXiv:2003.05672},
  year={2020}
}

@article{abbasimehr2022improving,
  title={Improving time series forecasting using LSTM and attention models},
  author={Abbasimehr, Hossein and Paki, Reza},
  journal={Journal of Ambient Intelligence and Humanized Computing},
  volume={13},
  number={1},
  pages={673--691},
  year={2022},
  publisher={Springer}
}

@article{holland2022marginal,
  title={Why marginal CO2 emissions are not decreasing for US electricity: estimates and implications for climate policy},
  author={Holland, Stephen P and Kotchen, Matthew J and Mansur, Erin T and Yates, Andrew J},
  journal={Proceedings of the National Academy of Sciences},
  volume={119},
  number={8},
  pages={e2116632119},
  year={2022},
  publisher={National Acad Sciences}
}

@article{Correlation,
  title={Long-Term Hourly Scenario Generation for Correlated Wind and Solar Power combining Variational Autoencoders with Radial Basis Function Kernels},
  author={Dias, Julio Alberto Silva},
  journal={arXiv preprint arXiv:2306.16427},
  year={2023}
}

@inproceedings{ashhab2024feasibility,
  title={Feasibility of Forecasting Highly Resolved Power Grid Frequency Utilizing Temporal Fusion Transformers},
  author={El Ashhab, Hadeer and Schäfer, Benjamin and Pütz, Sebastian},
  booktitle={NeurIPS 2024 Workshop on Tackling Climate Change with Machine Learning},
  url={https://www.climatechange.ai/papers/neurips2024/14},
  year={2024}
}

@inproceedings{effenberger2024towards,
  title={Towards turbine-location-aware multi-decadal wind power predictions with CMIP6},
  author={Effenberger, Nina and Ludwig, Nicole},
  booktitle={NeurIPS 2024 Workshop on Tackling Climate Change with Machine Learning},
  url={https://www.climatechange.ai/papers/neurips2024/32},
  year={2024}
}

@misc{C2ES2015,
  author = {{Center for Climate and Energy Solutions}},
  title = {EPA Regulation of Greenhouse Gas Emissions from New Power Plants},
  year = {2015},
  url = {https://www.c2es.org/document/epa-regulation-of-greenhouse-gas-emissions-from-new-power-plants/},
  note = {Accessed on [20,01,2025]}
}

@techreport{EEAGrants,
  author = {{EEA Grants}},
  title = {Conversion Guidelines},
  year = {2022},
  institution = {EEA Grants},
  type = {Technical Report},
  url = {https://www.eeagrants.gov.pt/media/2776/conversion-guidelines.pdf},
  note = {Accessed on [20,01,2025]}
}

\clearpage
\section{Appendices}

\subsection{ Day Ahead Economic Dispatch}
\label{DA}
\begin{align}
\label{eq:DA_obj}
Obj^{da} =Min \Big ( \sum_{v,t} C_v.p_{v,t} + \sum_{t}VOLL.ls_{t} \Big )
\end{align}
\begin{align}
\label{eq:DA_balance}
& \sum_v p_{v,t} + p^{rnw}_{t}+ls_{t}=P_{t} & \forall  t \\
&0 \leq p_{v,t} \leq \overline{P_v} & \forall v, t \\
&-R \leq p_{v,t} -p_{v,t-1} \leq R & \forall v, t \neq 1 \\
&-R \leq p_{v,t} -p_{v,T} \leq R & \forall v, t = 1\\
&0 \leq p_{t}^{rnw} \leq \overline{P^{rnw}} & \forall t \\
&0 \leq ls_{t}  \leq P_{t} & \forall t 
\end{align}

The DA objective function \ref{eq:DA_obj} aims to minimize both the generation cost and the load shedding cost. The equation \ref{eq:DA_balance} represents the supply-demand balance constraint, ensuring that total generation matches the demand at all times. Other equations impose constraints on the power system components, including upper and lower bounds on the power output of each generator, limits on renewable generation and load shedding, as well as restrictions on the ramping rates of generators. These ramping constraints specifically define how quickly generators can increase (ramp up) or decrease (ramp down) their power output. Collectively, these constraints ensure that the model accurately represents the physical and operational limitations of the power system components.

\subsection{Real Time Economic Dispatch}
\label{RT}
\begin{align}
Obj^{rt} = Min \Big (\sum_{v,t} C_v.p^{rt}_{v,t} + \sum_{t}VOLL.ls^{rt}_{t} \Big )
\end{align}
\begin{align}
& \sum_v p^{rt}_{v,t} + P^{rt,rnw}_{t} -p_{t}^{spill} -p^{rnw^{*}}_{t} +ls^{rt}_{t}=P_{t} & \forall t \\
&0 \leq p^{}_{v,t} 
 + p^{rt}_{v,t} \leq \overline{P_v} & \forall v\in V_f, t \\
&-R \leq (p^{*}_{v,t} -p^{*}_{v,t-1}) +p^{rt}_{v,t} -p^{rt}_{v,t-1} \leq R & \forall v \in V_f, t \neq 1 \\
&-R \leq(p^{*}_{v,t} -p^{*}_{v,t-T})+ p_{v^{rt},t} -p_{v^{rt},T} \leq R & \forall v \in V_f, t = 1\\
&0 \leq p_{t}^{spill} \leq P^{rnw,rt}_{t} & \forall  t \\
&0 \leq ls_{t} + ls^{rt}_{t}  \leq P_{t} & \forall  t 
\end{align}
These equations are similar to those used in day-ahead ED, but they aim to minimize costs in real-time. Variables marked with an asterisk (*) represent the optimal values determined in the day-ahead market.

\subsection{Characteristic of power system components }

\begin{table}[ht]
\centering
\begin{tabular}{cccccc}
\hline
Generator & \makecell{Generation \\ Cost (\$/MW)} & \makecell{Max \\ Generation \\ (MW)} & \makecell{Min \\ Generation \\ (MW)} & \makecell{ Availability \\at DA/RT}  &\makecell{Ramp \\up/down\\(MW/h)} \\
\hline
G1        & 20                   & 50                   & 0                   & DA     &20 \\
G2        & 25                   & 50                   & 0                   & DA     &20\\
G3        & 30                   & 30                   & 0                   & DA/RT  &30 \\
\hline
\end{tabular}
\caption{Characteristic of Generators}
\label{table:Generators}
\end{table}

\clearpage

\end{document}